\documentclass[runningheads]{llncs}
\usepackage{graphbox,graphicx}
\usepackage{amsmath,amssymb} 
\usepackage[colorlinks=true]{hyperref}
\usepackage{color}
\usepackage{url}
\usepackage[section]{placeins}
\usepackage[width=122mm,left=12mm,paperwidth=146mm,height=193mm,top=12mm,paperheight=217mm]{geometry}
\begin{document}

\pagestyle{headings}
\mainmatter

\title{Meaning guided video captioning}

\titlerunning{Meaning guided video captioning}

\authorrunning{Babariya and Tamaki}

\author{Rushi J. Babariya${}^1$, Toru Tamaki${}^2$
}
\institute{${}^1$BITS Pilani, India, ${}^2$Hiroshima University, Japan}

\maketitle

\begin{abstract}
Current video captioning approaches often
suffer from problems of missing objects in the video to be described, 
while generating captions semantically similar with ground truth sentences.
In this paper, we propose a new approach to video captioning that
can describe objects detected by object detection, and generate captions having similar meaning with correct captions. Our model relies on S2VT, a sequence-to-sequence model for video captioning. Given a sequence of video frames, the encoding RNN takes a frame as well as detected objects in the frame
in order to incorporate the information of the objects in the scene.
The following decoding RNN outputs are then fed into an attention layer and then to a decoder for generating captions. The caption is compared with the ground truth by learning metric so that vector representations of generated captions are semantically similar to those of ground truth.
Experimental results with the MSDV dataset demonstrate that
the performance of the proposed approach is much better than the model without the proposed meaning-guided framework, showing the effectiveness of the proposed model.
Code are publicly available at \url{https://github.com/captanlevi/Meaning-guided-video-captioning-}.

\end{abstract}

\section{Introduction}

The task of describing a video with a text has been receiving a great attention in recent years.
The mapping from a sequence of frames to a sequence of words was first introduced with a sequence-to-sequence model \cite{Venugopalan2015ICCV}, then a variety of models \cite{Aafaq2018survey} have been proposed.
However, these approaches often suffer from some common problems.
First, captions should reflect objects in the scene while generated captions may not include terms indicating such objects. This issue is caused by captioning models that take frames for capturing features, not for detecting objects in the scene.
Second, generated captions are evaluated with ground truth captions by using loss functions, which typically compare two sentences in a word-by-word manner. This may not reflect a semantic similarity between sentences because the change of a single word in a sentence could lead to a completely opposite meaning, but the loss might be small due to the small difference of the single word.

In this paper, we propose a meaning-guided video captioning model in cooperating with an object detection module.
Our model uses encoder LSTMs to learn the mapping from a video to a description,
of which back born network is 
the sequence-to-sequence video-to-text model, called S2VT \cite{Venugopalan2015ICCV}.
Upon this base network, we feed object detection results \cite{Redmon2018} of each frame into the encoder LSTMs
to extract the most dominant object in each frame.
Our model further incorporates attention in decoding LSTMs for enhancing information of frames that characterize the given video.
In addition, we proposes a new approach to train the proposed video-to-text model.
Instead of a classical training using a word-by-word loss,
we train the model to learn the meaning of captions, or semantic similarity of captions.
To this end, we propose a metric leaning model to embed captions so that
distances between a semantically similar pair of captions becomes smaller than a dissimilar pair.

\section{Related work}

There are many works on video captioning.
The early model was a sequence-to-sequence model (S2VT) \cite{Venugopalan2015ICCV}.
This was inspired by a sequence-to-sequence translation model
that takes a text in one language and output a text in another language.
Instead, the S2VT model takes a sequence of video frames as input to encoder LSTMs,
and outputs a sequence of words through decoder LSTMS.
Later a 3DCNN was used to extract video features \cite{Aafaq2018survey} to generate texts describing videos,
and also attention has been used \cite{Ashish2017NIPS} to find which part of the video are more informative.

Image captioning \cite{BAI2018NC,Hossain2019CSD,Liu2019VC} is a closely related task describing images, instead of videos.
Some works for image captioning
have been aware of the issue --- generated captions may miss objects in the scene \cite{Cornia2019CVPR} --- 
however not well studied for the video captioning task.
This could be alleviated by the help of object detection \cite{Redmon2018}.
We therefore use object detectors to find objects in the scene
and then reflect the object information in generated captions.

Designing the loss function is a key for many captioning models to success,
and for video captioning we need a loss to compare generated and ground truth captions.
This is common for many text-related tasks such as image and video captioning and visual question generation (VGQ) \cite{Li2018CVPR}.
A problem is that a loss usually compares texts word-by-word,
which is fragile to a little difference of words in sentences.
Furthermore, a typical dataset for captioning has several different captions as ground truth of a single video,
which is another cause for the word-by-word loss to be confused.
In the proposed model, we propose a loss using sentence embedding and metric leaning so that semantically similar captions have
small distances while different captions are far apart from each other.

\section{Encoder-decoder model}


\begin{figure}
  \centering
  \includegraphics[width=\linewidth]{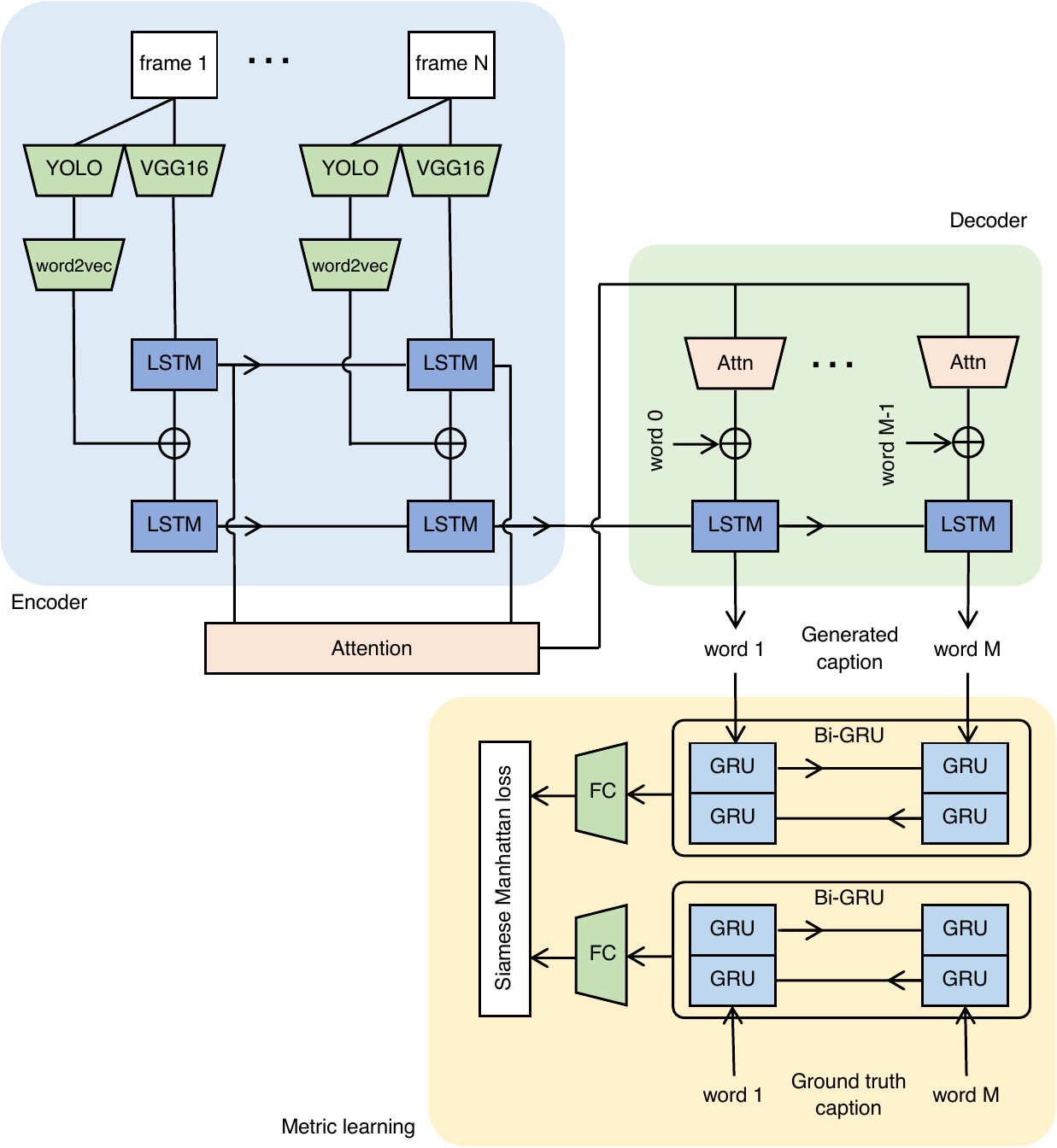}
  \caption{Our model}
  \label{fig:Our Model}
\end{figure}

Our proposed model is built on top of a baseline sequence-to-sequence video-to-text generator, S2VT \cite{Venugopalan2015ICCV}.
The baseline model uses a stacked 2-layer LSTM encoder-decoder model
that takes a sequence of RGB frames $f_1, f_2, \ldots, f_N$ as input and produces a caption or a sequence of words $w_1, w_2, \ldots, w_M$.
Frame features are extracted by using the VGG16 pre-trained model.
The lower LSTM in the encoder takes the output of the upper LSTM, encoding the visual information,
concatenated with padding due to the absence of text information.
In contrast, in the decoder the upper LSTM is fed padding due to the lack of video frames,
and the lower LSTM takes the concatenation of padding and word information.

The proposed model is shown in Figure \ref{fig:Our Model}.
It consists of encoder, decoder, and metric learning components.

\subsection{Encoder}
The encoder LSMT now takes the concatenation of holistic visual information 
and scene object information (instead of padding).
VGG16 features of 2048 dimension is extracted from a current RGB frame
as holistic visual information.
However it would not reflect objects in frames,
and therefore we use the YOLOv3 object detector \cite{Redmon2018}.
It may find many objects in a frame, however
we focus on the dominant object in each frame.
Specifically, we pick up the object having the highest objectness score in the
YOLO detector, and find the string describing the category of the object
(e.g., 'person' or 'cat').
The string is embedded with word2vec \cite{Mikolov2013NIPS,Mikolov2013ICLR},
pre-trained on a part of Google News Dataset \footnote{\url{https://code.google.com/archive/p/word2vec/}},
to convert it to an embedding vector of 300 dimension.

The upper LSTM in the encoder takes the 2048-d visual vector of the frame,
then the hidden state of 1000 dimension is passed to the lower LSTM
after concatenating with the object embedding vector,
resulting in a 1300-d vector to be fed to the lower LSTM.

The lower LSTM outputs a 1000-d hidden state vector that is passed through the encoder and to the decoder LSTM.
In contrast, 1000-d hidden states of the upper LSTM are
not passed to the decoder, but to the attention layer.

\subsection{Attention}
Let $h_1, \ldots, h_N$ be the hidden states of the upper encoder LSTM.
These are stacked in column-wise to make a matrix
\begin{equation}
    H = (h_1, \ldots, h_N) \quad \in R^{1000 \times N},
\end{equation}
where $N$ is the number of video frames encoded.
This is used as attention \cite{Ashish2017NIPS} for $s_t \in R^{1000}$, a given
output of the decoder LSTM at time step $t$ for $t=1,\ldots,M$.
To do so, we construct an activation energy vector of the following form \cite{Thang2015}
\begin{equation}
    \lambda_t = \operatorname{softmax}(s_t^T W H),
\end{equation}
where $W \in R^{1000 \times 1000}$ is a trainable linear layer.
Using $\lambda_t$, we have the attention vector of 1000 dimension
as $a_t = H \lambda_t$.

This attention vector is used as input at time step $t$
to the decoder LSTM
after a linear layer keeping dimension and concatenation with
the word embedding $w_t$ of 300 dimension at time $t$.

\subsection{Decoder}
The decoder LSTM takes the 1000-d hidden state from the lower decoder LSTM,
and the input (the concat of attention $a_t$ and word embedding $w_t$).
The output is 1000 dimension and fed into a linear layer
to convert a vector of vocabulary size of 25231 words
(in the case of the experiments below).
This is then passed to a softmax layer
to obtain the word probability $p_t$ at time $t$.

\subsection{Word-by-word loss}
This probability $p_t$ is used to compute the cross entropy
with the word $w_t$ in the ground truth caption.
The sum of these word-wise cross entropy values for $t=1,\ldots,M$ is
used as a loss to train the network.

This loss has been typically used for train networks
to compare generated and ground truth captions in the literature \cite{Vinyals2015CVPR,Li2018CVPR}.
However, it compares texts word-by-word,
which is fragile against a little difference of words in sentences.
Furthermore, a typical dataset for captioning has several different captions
as ground truth of a single video,
which is another cause for this word-by-word loss to be confused.

Therefore, in the training procedure,
we first use this loss to train the network until convergence,
then switch to another loss that captures the semantic similarity
between captions, which is described next.

\section{Metric learning component for captions}

\subsection{Soft-embedding sequence generation}

In order to construct a loss comparing generated and ground truth captions,
our model generates captions during training.
To this end, a possible way might be sampling the next word
by using the word probability $p_t$.
This is however not useful for training because the sampling
procedure cuts the computation graph and back propagation
doesn't go back through the decoder LSTM.

Instead, we propose to use the probability $p_t$
as weights for the next word.
If it was a one-hot vector, then
finding the next word is simply picking up the corresponding
column of the $300 \times 25231$ word embedding matrix $E$,
or equivalently multiplying the one-hot vector to $E$.
As the similar way, we construct a single word embedding
by $s_t = E p_t$.
This is actually not any of words in the vocabulary,
but should reflect a ``soft'' word choice of the decoder LSTM.

This is passed to the attention in the next time step,
then the next weighted embedding word $s_{t+1}$ is computed.
Eventually, the decoder LSTM outputs the sequence $s_1, \ldots, s_M$
as a generated caption for the given video.

\subsection{Meaning-guided loss}

Now we have two sequences; generated and ground truth captions.
In our metric learning component,
these captions are 
first embedded with a sentence-to-vector model.
This is a bi-directional GRU \cite{Schuster1997TSP} with 1000-d hidden states each,
resulting in 2000-d output.
Then a linear layer is used to reduce the dimension to 1000.

To compare two 1000-d vectors $v_1$ and $v_2$,
corresponding to generated and ground truth captions,
we use the the Siamese Manhattan loss \cite{Mueller2016SRA}
\begin{equation}
    L_{\mathrm{sim}}(v_1, v_2) = 1 - \exp( \| v_1 - v_2 \|_1 ).
\end{equation}
This loss should be small for $v_1$ and $v_2$ 
because these two captions should be similar and the vectors should also be close to each other.
This assumes that the model generates a reasonable caption $v_1$
that should be semantically similar to $v_2$.
However in the early stage of the training,
the model might be giving a very different caption from the ground truth,
and if this is the case then the network might learn an identity mapping
where it says that any caption pairs be semantically similar.
To prevent this,
we also use two different captions $v_3$ and $v_4$,
and minimize the following loss as well;
\begin{equation}
    L_{\mathrm{dis}}(v_3, v_4) = \exp( \| v_3 - v_4 \|_1 ).
\end{equation}
The overall loss for this is given by
\begin{equation}
\begin{aligned}
L 
=& E_{v_1, v_2 \sim \text{training sample pair}}[L_{\mathrm{sim}}(v_1, v_2)] + \\
 & E_{v_3, v_4 \sim \text{dissimilar sentence pair}}[L_{\mathrm{dis}}(v_3, v_4)].
\end{aligned}
\end{equation}
We need to pre-train the model for this to work,
as we will describe later.


\subsection{Intra-batch training}

A possible drawback of the metric leaning component described above
is that datasets for captioning do not provide any dissimilar sentence pairs.
In the followings, we describe tricks to train the proposed model efficiently.

\begin{figure}[t]
  \centering
  \includegraphics[width=.7\linewidth]{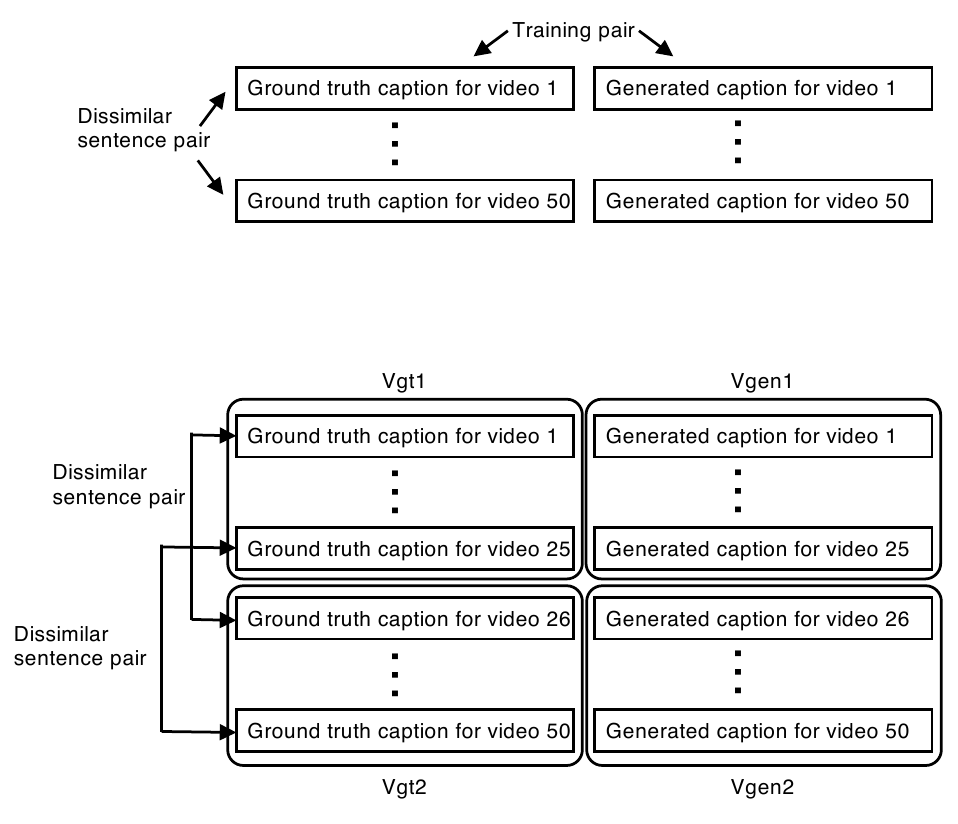}
  \caption{Intra-batch sampling.
  (top) Sampling two ground truth sentences in a batch
  as dissimilar sentence pairs.
  (bottom) Use corresponding ground truth pairs as 
  as dissimilar sentence pairs.
  }
  \label{fig:intra-batch sampling}
\end{figure}

The first trick is to use a mini-batch for dissimilar pair sampling
(Figure \ref{fig:intra-batch sampling}(top)).
Suppose we are given a batch consists of 50 ground truth captions
for 50 different videos in a dataset, and then we have corresponding
50 generated captions.
Among these 100 captions, the 50 training pairs are used for
$L_{\mathrm{sim}}$. In addition, there are many more
dissimilar caption pairs because different ground truth captions
can be considered as different sentences.
Therefore we can sample different caption pairs inside the batch for
$L_{\mathrm{dis}}$.

However, a naive sampling is inefficient because 
a ground truth caption is encoded in a vector several times;
once as a training (similar) pair, and more as a dissimilar pair.
This leads to encoding the same caption multiple times 
with the the network having the same weights.
This is a waste of resource because 
the encoding results are the same before computing the loss and backprop.

Our second trick is to do this efficiently
(Figure \ref{fig:intra-batch sampling}(bottom)).
Again suppose we are given a 50-sample batch,
and we have generated and ground truth caption embeddings
as two matrices of size $50 \times 1000$
(each row is a 1000-d embedding vector).
Let $V_{\mathrm{gen}}$ is the matrix of generated captions,
and $V_{\mathrm{gt}}$ is the matrix of ground truth captions.
We split them to two to obtain
four $25 \times 1000$ matrices; 
$V_{\mathrm{gen1}}, V_{\mathrm{gt1}}$ and
$V_{\mathrm{gen2}}, V_{\mathrm{gt2}}$.
Now each row in $V_{\mathrm{gen1}}$ has nothing related to 
any row in $V_{\mathrm{gen2}}$ due to random sampling from the training dataset,
and the same for $V_{\mathrm{gt1,2}}$.
Therefore, we can use $i$-th rows of 
$V_{\mathrm{gen2}}, V_{\mathrm{gt2}}$ as the dissimilar pair
for $i$-th rows of $V_{\mathrm{gen1}}, V_{\mathrm{gt1}}$, and vice versa.
This can be done by keeping these embedding vectors just before
computing the Siamese Manhattan loss.

Our third trick is to do it more efficiently.
As a concept, $i$-th training pair and $i$-th dissimilar pair
are used for computing the loss, for $i=1,\ldots$ over training samples in the batch.
However, usually the loss is aggregated for training samples in the batch,
to compute the loss value of the batch.
Here the order doesn't matter; we can compute and aggregate
the loss $L_{\mathrm{sim}}$ for training samples first.
Then we can add the loss $L_{\mathrm{dis}}$ to compute the final loss value of the batch.

The final trick is to use two different optimizers.
For computing the loss $L_{\mathrm{sim}}$ for training samples,
video frames are input the network to generate a caption.
However, 
the loss $L_{\mathrm{dis}}$ is only used for learning the metric learning component,
not for the encoder-decoder LSTMs.
In other words, only $v_1$ connects the loss and the encoder and decoder components;
$v_2$ is a given ground truth, and $v_3, v_4$ are only used for the metric learning component.
Therefore we use different optimizers (updaters) for two losses.
For training with the loss $L_{\mathrm{sim}}$, one optimizer updates all networks weights.
For training with the loss $L_{\mathrm{dis}}$, another optimizer updates weights in the metric learning component only.

\section{Experiments}

\subsection{Dataset}

The dataset used is the Microsoft Video Description corpus (MSVD) \cite{Chen2011CHP}.
This is a collection of Youtube clips (1970 in total), average length of videos is about 6 seconds.
Each video has descriptions given by several annotators who describe the video in one sentence
(40 captions per video on average).
The data is split in the following way \cite{Venugopalan2015ICCV};
1200 videos are used for training, and 100 for validation.
The remaining 670 are used for testing.
For a single video, we used up to $N=80$ frames as input to the encoder LSTM.


\subsection{Training procedure}

The metric learning component
uses the Siamese Manhattan loss,
however we use a triplet loss
for pre-training the metric learning component.
Here we can use the similar trick
explained in the section of intra-batch training.
A triplet loss takes three arguments; reference,
positive, and negative samples.
Given a batch of 50 samples,
we have 50 pairs of generated and ground truth
captions. For one of these pairs,
the other 49 samples can be considered 
as negatives for the triplet loss.
This is much more efficient than a naive training.

For the encoder and decoder components,
we use the word-by-word loss
without the metric learning component.
Once the pre-training phase has been done,
then we use the both losses for training
in a stochastic manner.
Specifically, given a batch,
we randomly select if the metric leaning
component is used (and then the Siamese Manhattan loss)
in chance of 70\%,
or not (the word-by-word loss is used) in 30\%.

\subsection{Results}

Table \ref{tab:results} shows
results of the baseline and proposed models.
There are three different settings for the proposed models;
O stands for the case using object information only (hence the attention and metric leaning component are not used),
OA stands for the case when object information and attention are used but not the metric learning component,
and OAM stands for the full model including
the object, attention, and metric learning components.

The use of the object information
clearly improve the performance
against the baseline (original sequence-to-sequence)
model. This is because main objects in each frame
are explicitly used in the encoder.
Attention further improves the results (except CIDEr),
which is expected as many results reported with
better performance with attention.
Our full model, shown as OAM in Tab. \ref{tab:results},
can further boost the performance by 1.1\% in BLUE4.
This is not as large as improvements with object information (by 10.7\%)
and attention (2.9\%), but this results suggest
that the proposed metric learning component
can be used to improve results by adding to any models
other than the sequence-to-sequence architecture.

\begin{table}[t]
    \centering
    \caption{Results
    of baseline and proposed models on the MSVD dataset.
      Symbols stand for;
  O -- Objects,
  A -- Attention,
  M -- Meaning model.
    }
    \label{tab:results}
\begin{tabular}{|l|c|c|c|}
\hline
Model & BLEU4 & METEOR & CIDEr  \\
\hline
Baseline (S2VT) 
    & 0.288 & 0.246 & --  \\
O   & 0.395 & 0.295 & 0.641 \\
OA  & 0.424 & 0.312 & 0.641 \\
OAM & 0.435 & 0.316 & 0.649 \\
\hline
\end{tabular}
\end{table}

\begin{figure}[t]
\includegraphics[width=3cm,align=c]{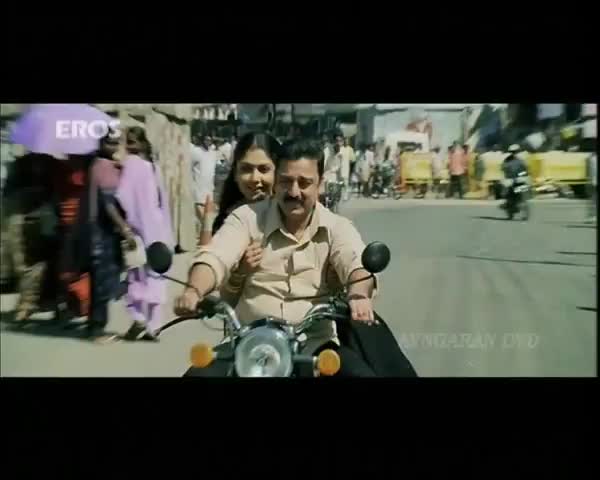}
\hspace{1em}
\begin{tabular}{|l|l|}\hline
GT  & A man and woman ride a motorcycle. \\ \hline
O   & a man is dancing. \\
OA  & two people are dancing.   \\
OAM & a man and a woman are riding a motorcycle. \\ \hline
\end{tabular}

\smallskip

\includegraphics[width=3cm,align=c]{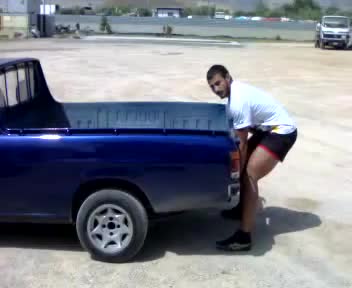}
\hspace{1em}
\begin{tabular}{|l|l|}\hline
GT  & A man is lifting the car. \\ \hline
O   & a man is lifting a car. \\
OA  & a man is lifting the back of a truck. \\
OAM & a man is lifting a truck. \\ \hline
\end{tabular}

\smallskip

\includegraphics[width=3cm,align=c]{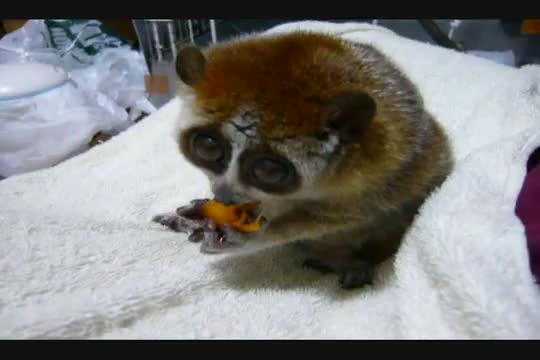}
\hspace{1em}
\begin{tabular}{|l|l|}\hline
GT  & An animal is eating. \\ \hline
O   & a cat is licking a lollipop. \\
OA  & a dog is eating. \\ 
OAM & a dog is eating. \\ \hline
\end{tabular}

  \caption{Examples of generated captions.
  Left images are frames of videos,
  and right texts are ground truth and 
  generated captions.
  Symbols stand for;
  O -- Objects,
  A -- Attention,
  M -- Meaning model.
  }
  \label{fig:generatd captions}
 \end{figure}

Figure \ref{fig:generatd captions} shows
some examples generated by the proposed model.
In the first video (top row in the figure),
the scene changes frequently because of the movie editing; the video is
composed of several cuts from different angles.
Therefore the model without the metric learning component
is confused and generated ``dancing'' instead of ``riding''.

In the last video (bottom row of the figure),
the caption generated by the full model
is considered as wrong because the ground truth caption
mentions the animal as ``animal'' and it is obviously not a dog,
while the generated caption says that it is a ``dog''.
This is a limitation of the proposed model
because the mistakes of the object detector (``dog'' for ``animal'')
directly affect the encoder.

Table \ref{tab:results2} shows a comparison
with other recent methods.
While the proposed method doesn't perform
as like recent methods,
the proposed meaning-guided loss and tricks for intra-batch training
are expected to work for boosting other methods.

\begin{table}
    \centering
    \caption{Results
    of the proposed model and other methods.}
    \label{tab:results2}
\begin{tabular}{|c|c|c|c|}
\hline
Model & BLEU4 & METEOR & CIDEr  \\
\hline
OAM (ours) & 0.435 & 0.316 & 0.649 \\
\cite{Wang_2018_CVPR} & 0.523 & 0.341 & 0.698 \\
\cite{Aafaq_2019_CVPR} & 0.479 & 0.350 & 0.781 \\
\hline
\end{tabular}
\end{table}

\section{Conclusions}
We have proposed a model for video captioning
guided by the similarity between captions.
The proposed model has three components;
the 2-layer LSTM encoder involving scene object information,
the LSTM decoder with attention,
and the metric learning for comparing two captions in a latent space.
Experimental results show that the proposed model
outperforms the baseline, a sequence-to-sequence model,
by introducing the metric learning component
in addition to attention mechanism and object information.

Our future work includes using the proposed metric learning component
in other state-of-the-art caption models where word-by-word losses are used \cite{Wang_2018_CVPR,Aafaq_2019_CVPR},
and incorporating other corpuses for sentence-level similarity computation,
such as \cite{Bowman2015emnlp}.

\section*{Acknowledgement}

This work was supported by 
International Linkage Degree Program (ILDP) in Hiroshima University (HU).
This work was also supported by JSPS KAKENHI grant number JP16H06540.

\bibliographystyle{splncs}
\bibliography{egbib}

\begin{thebibliography}{10}

\bibitem{Venugopalan2015ICCV}
{Venugopalan}, S., {Rohrbach}, M., {Donahue}, J., {Mooney}, R., {Darrell}, T.,
  {Saenko}, K.:
\newblock Sequence to sequence -- video to text.
\newblock In: 2015 IEEE International Conference on Computer Vision (ICCV).
  (Dec 2015)  4534--4542

\bibitem{Aafaq2018survey}
Aafaq, N., Gilani, S.Z., Liu, W., Mian, A.:
\newblock Video description: {A} survey of methods, datasets and evaluation
  metrics.
\newblock CoRR \textbf{abs/1806.00186} (2018)

\bibitem{Redmon2018}
Redmon, J., Farhadi, A.:
\newblock Yolov3: An incremental improvement.
\newblock CoRR \textbf{abs/1804.02767} (2018)

\bibitem{Ashish2017NIPS}
Vaswani, A., Shazeer, N., Parmar, N., Uszkoreit, J., Jones, L., Gomez, A.N.,
  Kaiser, L.u., Polosukhin, I.:
\newblock Attention is all you need.
\newblock In Guyon, I., Luxburg, U.V., Bengio, S., Wallach, H., Fergus, R.,
  Vishwanathan, S., Garnett, R., eds.: Advances in Neural Information
  Processing Systems 30.
\newblock Curran Associates, Inc. (2017)  5998--6008

\bibitem{BAI2018NC}
Bai, S., An, S.:
\newblock A survey on automatic image caption generation.
\newblock Neurocomputing \textbf{311} (2018)  291 -- 304

\bibitem{Hossain2019CSD}
Hossain, M.Z., Sohel, F., Shiratuddin, M.F., Laga, H.:
\newblock A comprehensive survey of deep learning for image captioning.
\newblock ACM Comput. Surv. \textbf{51}(6) (February 2019)  118:1--118:36

\bibitem{Liu2019VC}
Liu, X., Xu, Q., Wang, N.:
\newblock A survey on deep neural network-based image captioning.
\newblock The Visual Computer \textbf{35}(3) (Mar 2019)  445--470

\bibitem{Cornia2019CVPR}
Cornia, M., Baraldi, L., Cucchiara, R.:
\newblock Show, control and tell: A framework for generating controllable and
  grounded captions.
\newblock In: The IEEE Conference on Computer Vision and Pattern Recognition
  (CVPR). (Jun 2019)  8307--8316

\bibitem{Li2018CVPR}
Li, Y., Duan, N., Zhou, B., Chu, X., Ouyang, W., Wang, X., Zhou, M.:
\newblock Visual question generation as dual task of visual question answering.
\newblock In: The IEEE Conference on Computer Vision and Pattern Recognition
  (CVPR). (April 2018)  6116--6124

\bibitem{Mikolov2013NIPS}
Mikolov, T., Sutskever, I., Chen, K., Corrado, G., Dean, J.:
\newblock Distributed representations of words and phrases and their
  compositionality.
\newblock In: Proceedings of the 26th International Conference on Neural
  Information Processing Systems - Volume 2. NIPS'13, USA, Curran Associates
  Inc. (2013)  3111--3119

\bibitem{Mikolov2013ICLR}
Mikolov, T., Chen, K., Corrado, G., Dean, J.:
\newblock Efficient estimation of word representations in vector space.
\newblock In: 1st International Conference on Learning Representations, {ICLR}
  2013, Scottsdale, Arizona, USA, May 2-4, 2013, Workshop Track Proceedings.
  (2013)

\bibitem{Thang2015}
Luong, T., Pham, H., Manning, C.D.:
\newblock Effective approaches to attention-based neural machine translation.
\newblock In: Proceedings of the 2015 Conference on Empirical Methods in
  Natural Language Processing, Lisbon, Portugal, Association for Computational
  Linguistics (September 2015)  1412--1421

\bibitem{Vinyals2015CVPR}
Vinyals, O., Toshev, A., Bengio, S., Erhan, D.:
\newblock Show and tell: A neural image caption generator.
\newblock In: The IEEE Conference on Computer Vision and Pattern Recognition
  (CVPR). (June 2015)

\bibitem{Schuster1997TSP}
{Schuster}, M., {Paliwal}, K.K.:
\newblock Bidirectional recurrent neural networks.
\newblock IEEE Transactions on Signal Processing \textbf{45}(11) (Nov 1997)
  2673--2681

\bibitem{Mueller2016SRA}
Mueller, J., Thyagarajan, A.:
\newblock Siamese recurrent architectures for learning sentence similarity.
\newblock In: Proceedings of the Thirtieth AAAI Conference on Artificial
  Intelligence. AAAI'16, AAAI Press (2016)  2786--2792

\bibitem{Chen2011CHP}
Chen, D.L., Dolan, W.B.:
\newblock Collecting highly parallel data for paraphrase evaluation.
\newblock In: Proceedings of the 49th Annual Meeting of the Association for
  Computational Linguistics: Human Language Technologies - Volume 1. HLT '11,
  Stroudsburg, PA, USA, Association for Computational Linguistics (2011)
  190--200

\bibitem{Wang_2018_CVPR}
Wang, B., Ma, L., Zhang, W., Liu, W.:
\newblock Reconstruction network for video captioning.
\newblock In: The IEEE Conference on Computer Vision and Pattern Recognition
  (CVPR). (June 2018)

\bibitem{Aafaq_2019_CVPR}
Aafaq, N., Akhtar, N., Liu, W., Gilani, S.Z., Mian, A.:
\newblock Spatio-temporal dynamics and semantic attribute enriched visual
  encoding for video captioning.
\newblock In: The IEEE Conference on Computer Vision and Pattern Recognition
  (CVPR). (June 2019)

\bibitem{Bowman2015emnlp}
Bowman, S.R., Angeli, G., Potts, C., Manning, C.D.:
\newblock A large annotated corpus for learning natural language inference.
\newblock In: Proceedings of the 2015 Conference on Empirical Methods in
  Natural Language Processing (EMNLP), Association for Computational
  Linguistics (2015)

\end{thebibliography}
\end{document}